\documentclass{article}
\usepackage[preprint]{corl_2026}
\hypersetup{
  pdfauthor={Immanuel R. Santjoko, Benedictus C. G. Cinun,
             Jing Chen, Bin Hu},
  pdfsubject={Robotics preprint}
}

\usepackage{url}
\usepackage{amsmath,amsfonts}
\usepackage{algorithm}
\usepackage{algorithmic}
\usepackage{array}
\usepackage{booktabs}
\usepackage{float}
\usepackage{graphicx}
\usepackage{multirow}
\usepackage{subcaption}
\usepackage{textcomp}
\usepackage{color}
\usepackage{wrapfig}
\usepackage{xspace}
\newcommand{\ours}{\textsc{RCRA}\xspace}

\title{Learning When to Adapt:\\ Reactivity-Conditioned Residual Adaptation\\ under Human--Robot Distribution Shift}

\author{%
  Immanuel R. Santjoko$^{1}$ \quad
  Benedictus C. G. Cinun$^{1}$ \quad
  Jing Chen$^{3}$ \quad
  Bin Hu$^{1,2}$ \\
  $^{1}$Department of Electrical \& Computer Engineering,
  University of Houston \\
  $^{2}$Department of Engineering Technology,
  University of Houston \\
  $^{3}$Department of Psychological Sciences,
  Rice University \\
  Houston, TX, USA
}

\begin{document}

\maketitle

\begin{abstract}
Pedestrians differ in whether they respond to a robot, yet many RL policies assume all are non-reactive to prioritize safety.
Policies trained without conditioning on local pedestrians' reactivity may therefore be unnecessarily conservative or rely on cooperation that does not occur.
We propose \ours, an online adaptation framework that modulates a learned residual around a frozen conservative navigation policy according to estimated pedestrian reactivity.
A Transformer-based estimator infers latent pedestrian reactivity from motion histories and robot context, providing a more accurate signal for navigation adaptation and improving scenario-averaged simulation accuracy by $12.62$ percentage points over the Bayesian baseline, which assumes known pedestrian goals.
These estimates are spatially aggregated into an adaptation scale that gates LoRA modules, enabling cooperative behavior near reactive pedestrians and conservative behavior near non-reactive pedestrians.
Among baselines achieving SR above $90\%$ in every scenario, \ours reduces scenario-averaged path length by $16.7\%$ relative to the conservative backbone and by $7.3\%$ relative to randomized-training baselines.
It improves mean SR by $14.88$ percentage points over the reactive-trained policy.
Website is available at
\url{https://nail-uh.github.io/rcra-website.github.io/}.
\end{abstract}



\section{Introduction}\label{sec:intro}

Pedestrians respond heterogeneously to mobile robots: some adjust their motion in response to the robot, while others continue along their nominal paths.
We study \emph{human--robot interaction distribution shift}: pedestrian responses to the robot can differ between training and deployment and vary across individuals, locations, and time.
Policies trained under non-reactive pedestrian assumptions tend to preserve safety, but they can remain overly conservative when pedestrians actually make space.
Conversely, policies adapted to fully cooperative pedestrians can move efficiently but may collide when deployed around pedestrians who do not yield.
Uncertainty-based navigation methods can respond to prediction mismatch by increasing safety margins~\citep{yao2024sonic,yao2025towards}.
However, in interaction-sensitive crowds, not all prediction errors indicate danger: a pedestrian who yields may deviate from a non-reactive forecast in a way that creates more free space for the robot.
Increasing margins alone can therefore overlook opportunities created by robot-directed responses. This motivates estimating how pedestrians respond to the robot and using that estimate to adjust navigation behavior.

We investigate whether explicit estimates of pedestrian reactivity provide a useful signal for modulating a pretrained navigation policy as local interaction conditions change.
We propose reactivity-conditioned residual adaptation (\ours), which couples two components: a learned estimator of latent robot-directed reactivity and a policy residual modulated by those estimates. The estimator uses human motion history and robot context to infer which pedestrians are likely to respond. Spatial aggregation converts these estimates into a continuous scale that modulates a residual learned from reactive crowds around a frozen conservative policy. This lets the robot adjust to local interactions without online gradient updates or a supplied scenario label.  
Our contributions are threefold. First, we introduce \ours for human--robot interaction distribution shift, using estimated pedestrian responses to modulate an interaction-specific residual around a frozen conservative navigation policy. Second, we learn per-pedestrian reactivity from partial observation histories, refine the estimator with DAgger, and use it for policy modulation through spatial relevance weights. Finally, we evaluate the empirical safety--efficiency trade-off across five homogeneous, heterogeneous, and time-varying response regimes, and separately validate residual adaptation in controlled robot trials with prescribed reactivity labels.

\section{Related Work}\label{sec:related}

\paragraph{Crowd Navigation and Prediction Uncertainty.}
Classical crowd-navigation methods use structured motion and collision-avoidance models~\citep{helbing1995social,fiorini1998motion,van2011reciprocal}, while cooperative planners account for human responses to robot motion~\citep{trautman2010unfreezing,trautman2015robot}.
Learning-based methods represent crowd interactions through reinforcement learning and attention~\citep{chen2017socially,everett2018motion,chen2019crowd,liu2021decentralized}, with trajectory prediction supporting anticipatory navigation~\citep{liu2023intention}.
Uncertainty-aware methods incorporate safety costs and adaptive prediction bounds~\citep{yao2025towards,huang2025interaction}, although standard conformal guarantees require assumptions such as exchangeability~\citep{angelopoulos2021gentle}.
Prediction mismatch alone does not distinguish hazardous motion from yielding that creates free space, motivating explicit estimation of pedestrian responses to the robot.

\paragraph{Reactivity Estimation and Policy Adaptation.}
Training across varied pedestrian-response conditions can improve robustness~\citep{shen2026human}, but does not explicitly condition online adaptation on estimated robot-directed reactivity.
BDDS uses Bayesian inference to estimate human responses~\citep{hosseinzadeh2021toward}, but its behavioral cost model requires pedestrian goals, which may be unknown or change during deployment.
Policy adaptation also includes domain randomization, meta-learning, and imitation-learning correction~\citep{tobin2017domain,finn2017model,ross2011reduction}.
Parameter-efficient methods adapt pretrained networks while freezing most parameters~\citep{houlsby2019parameter,hu2022lora,liu2024tail}.
Context-dependent adapter selection and scaling further enable dynamic modulation of pretrained models~\citep{schmied2023learning}.
We combine goal-free reactivity estimation with online modulation of a learned LoRA residual around a frozen conservative policy, adjusting navigation to local responses without online gradient updates.

\section{Reactivity-Conditioned Residual Adaptation}\label{sec:method}
\paragraph{System Framework and Problem Formulation.}\label{sec:problem_formulation}
We formulate crowd navigation as a partially observable Markov decision process (POMDP), because robot-directed reactivity is not directly observed. Let $x_t^r$ denote the robot state at timestep $t$, including its position, velocity, heading, and goal. For pedestrian $i$, let $x_{i,t}^h=(p_{i,t}^h,v_{i,t}^h,z_{i,t})$, where $p_{i,t}^h$ and $v_{i,t}^h$ are planar position and velocity, and $z_{i,t}\in\{0,1\}$ is a latent indicator of robot-directed reactivity. 
Here, $z_{i,t}=1$ denotes a reactive pedestrian whose motion accounts for the robot, whereas $z_{i,t}=0$ denotes a non-reactive pedestrian whose motion does not.
We study interaction shifts in the distribution and evolution of these modes, which alter pedestrian responses to the robot.
We write the state compactly as $s_t=(x_t^r,\{x_{i,t}^h\}_{i=1}^{N})$, with pedestrian goals and fixed agent attributes implicit in this notation. $N$ is the total number of pedestrians. The robot observes its own state and tracks $N_t\leq N$ pedestrians within sensing range, but does not observe their reactivity labels. We construct the augmented observation
$o_t = \left(x_t^r,\{p_{i,t}^h\}_{i=1}^{N_t},
 \{\hat\tau_{i,t+1:t+K\mid t}\}_{i=1}^{N_t},
 \{\hat\delta_{i,t+1:t+K\mid t}\}_{i=1}^{N_t}\right),$
where $\hat\tau_{i,t+1:t+K\mid t}$ is the trajectory predicted at time $t$~\citep{huang2021learning}, $\hat\delta_{i,t+1:t+K\mid t}$ denotes the associated prediction-uncertainty features used by GenSafeNav~\citep{yao2025towards}, and $K$ is the prediction horizon in timesteps. Position histories provide motion information, with pedestrian identities maintained by tracking. The robot's belief about reactivity is
$b_{i,t}=P(z_{i,t}=1\mid o_{1:t}).$
Here, $o_{1:t}$ denotes the observation history, and $b_{i,t}\in[0,1]$ represents uncertainty about a binary response mode, not a degree of reactivity.

\begin{figure}[t]
 \centering
 \includegraphics[width=\linewidth]{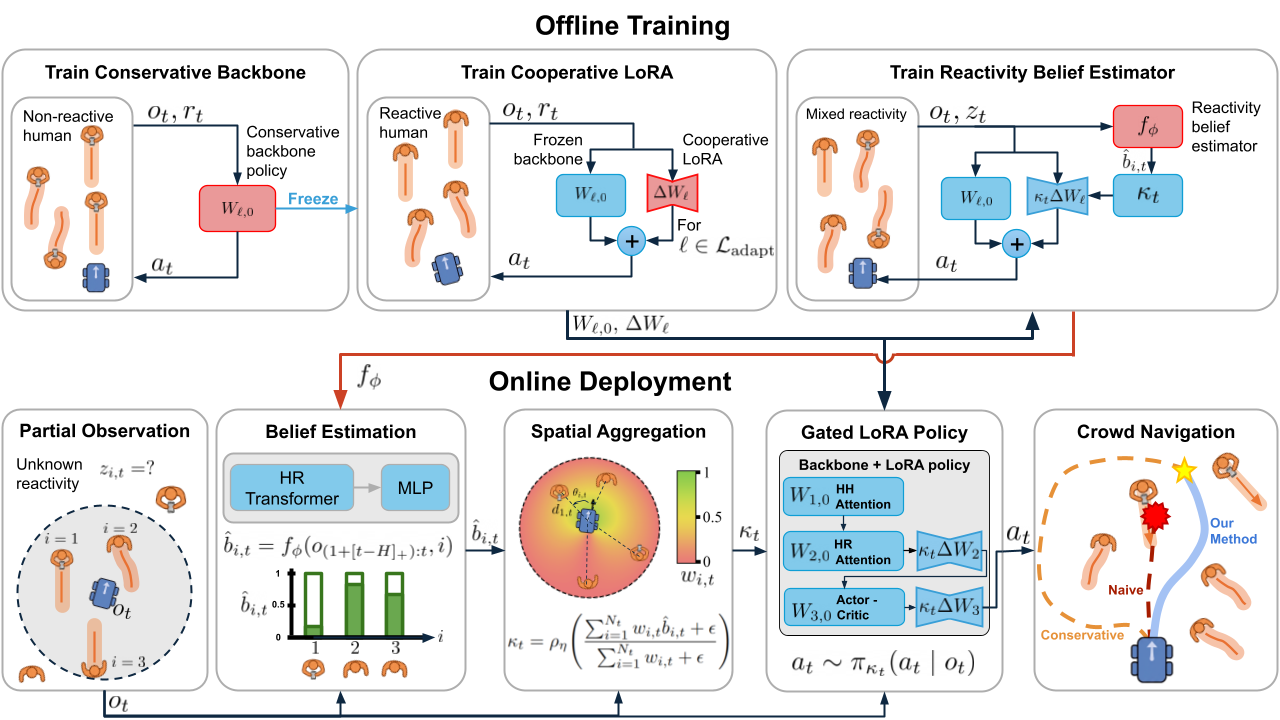}
 \caption{\textbf{Overview of \ours.} Offline, a conservative backbone is trained on non-reactive pedestrians and then frozen while LoRA residual modules are learned for reactive-crowd behavior. A reactivity estimator is trained on mixed scenarios. Online, the robot estimates per-human reactivity beliefs from partial observations, aggregates them using spatial relevance weights, and uses the resulting scale $\kappa_t$ to gate the LoRA residual to adapt the navigation safety–efficiency operating point.}
 \label{fig:method_overview}
\end{figure}

This setting presents two coupled tasks: infer latent reactivity from partial histories and use those estimates to adjust navigation under changing interaction dynamics. As illustrated in Fig.~\ref{fig:method_overview}, \ours combines a residual policy family, a learned belief estimator, and spatial aggregation into the continuous adaptation scale $\kappa_t$. The navigation objective and training parameters follow GenSafeNav (Appendix~\ref{sec:implementation_details}). We first describe the policy family, then the estimator and the rule connecting them.

\paragraph{A Reactivity-Conditioned Policy.}
We represent navigation behavior using a frozen conservative backbone and a learned interaction-specific parameter residual. We first train the backbone on non-reactive crowds and freeze its parameters $\theta_0$. We then learn a residual $\Delta\theta$ from reactive-crowd rollouts while keeping $\theta_0$ fixed.
We parameterize the residual with LoRA.
For an adapted linear layer $\ell$ with frozen weight $W_{\ell,0}\in\mathbb{R}^{m\times n}$, the learnable update is $\Delta W_\ell=A_\ell B_\ell$,
where $A_\ell\in\mathbb{R}^{m\times r}$, $B_\ell\in\mathbb{R}^{r\times n}$, and $r\ll\min(m,n)$.
Here, $m$ and $n$ are the layer's output and input dimensions, and $r$ is the adaptation rank.
The fixed LoRA gain is absorbed into these factors.
We adopt the GenSafeNav architecture~\citep{yao2025towards}, which combines human--human attention, human--robot attention, a GRU, and actor--critic heads.
We inject LoRA only into the human--robot attention layers and actor--critic heads, since latent reactivity primarily affects how the robot interprets human--robot interaction and selects actions.
Human--human attention remains frozen to preserve the learned crowd-interaction representation.
Thus, $\Delta\theta$ contains the low-rank updates for these adapted layers and zeros for all remaining parameters.
Appendix~\ref{sec:adaptation_mechanism_ablation} compares the computational costs of this implementation with gated full fine-tuning and action-space interpolation.

At runtime, a reactivity-dependent scale $\kappa_t\in[0,1]$ controls the residual strength, giving the active parameters
 $ \theta_t=\theta_0+\kappa_t\Delta\theta.$
Each adapted layer uses $W_{\ell,t}=W_{\ell,0}+\kappa_t A_\ell B_\ell$ for $\ell\in\mathcal L_{\mathrm{adapt}}$, while all other layers retain $W_{\ell,t}=W_{\ell,0}$. Here, $\mathcal L_{\mathrm{adapt}}$ denotes the set of LoRA-adapted layers. This yields the policy family
\begin{equation}
 \pi_{\kappa_t}(a_t\mid o_t)
 =\pi(a_t\mid o_t;\theta_0+\kappa_t\Delta\theta),
 \label{eq:policy_family}
\end{equation}
where $a_t$ is the robot's velocity command. We suppress the policy's GRU hidden state in this notation. 
The policy also depends on its recurrent history. Setting $\kappa_t=0$ disables the residual and uses the conservative parameters, while $\kappa_t=1$ activates the full learned residual. Intermediate values continuously interpolate the parameters.
They do not guarantee monotonic changes in actions or safety. We obtain $\kappa_t$ from estimated per-pedestrian beliefs and spatial relevance, as described next.

\paragraph{Reactivity Belief Estimation.}\label{sec:reactivity_belief_estimation}
The inference problem is to estimate the reactivity belief $b_{i,t}$ from a finite history of the robot's observations.
We estimate this using a learned reactivity estimator
\begin{equation}\label{eq:estimate}
 \hat b_{i,t}=f_\phi(o_{(1+[t-H]_+):t},i)\in[0,1],
\end{equation}
where $[x]_+=\max(0,x)$, $H$ is the history length in timesteps,
and $i$ identifies the pedestrian track.
This window provides temporal context for identifying motion adjustments in response to the robot while limiting the influence of outdated behavior.
To process observation histories, each pedestrian's augmented state and the robot state are embedded with MLPs at each timestep.
A multi-head Transformer encoder with padding masks captures interactions among nearby pedestrians, and an MLP outputs $\hat b_{i,t}$ for each observed pedestrian.
We train $f_{\phi}$ using the current simulator labels $z_{i,t}$ with binary cross-entropy and refine it with DAgger~\citep{ross2011reduction} to reduce closed-loop covariate shift.
At each DAgger-style aggregation iteration, we roll out the current adaptive policy, label visited observations using the simulator's reactivity labels, and update $f_\phi$ on the aggregated data. The policy backbone and learned residual remain fixed. This addresses the estimator's policy-induced covariate shift, which is distinct from the changing pedestrian response dynamics targeted by \ours.

\paragraph{Spatial Reactivity Aggregation.}\label{sec:spatial_belief_aggregation}
A global average of $\hat{b}_{i,t}$ may obscure locally important differences in mixed crowds: distant reactive pedestrians can dilute the influence of a nearby non-reactive pedestrian.
We therefore weight reactivity beliefs by distance and bearing relative to the robot's heading, emphasizing nearby pedestrians ahead of the robot when determining the residual strength.
This spatial aggregation provides a tractable control heuristic rather than a solution to the full POMDP.
For each observed pedestrian, we define a spatial relevance weight
\begin{equation}
w_{i,t} = \left[ 1 - \frac{d_{i,t}}{R} \right]_+ \left[\cos\left(\frac{\psi_{i,t}}{2}\right)\right]_+,
 \label{eq:spatial_weight_revised}
\end{equation}
where $R>0$ is the sensing radius, $d_{i,t}$ is the robot--pedestrian distance, and $\psi_{i,t}\in[-\pi,\pi]$ is the pedestrian's bearing relative to the robot's heading. These weights approximate spatial relevance to near-term navigation.
The runtime adaptation scale is then computed as
\begin{equation}
\kappa_t =
\rho_\eta\!\left(
\frac{\sum_{i=1}^{N_t} w_{i,t}\hat b_{i,t}+\epsilon}
{\sum_{i=1}^{N_t} w_{i,t}+\epsilon}
\right),
\qquad
\rho_\eta(\bar{\kappa})=
\frac{\eta\bar{\kappa}}
{\eta\bar{\kappa}+(1-\eta)(1-\bar{\kappa})},
\quad \kappa_t,\bar{\kappa}\in[0,1].
\label{eq:adaptive_scale}
\end{equation}
where $\epsilon>0$ prevents division by zero and biases the aggregation toward full residual activation when total spatial relevance is small.
The bias-control parameter $\eta\in(0,1)$ adjusts the safety--efficiency preference:
$\eta=0.5$ preserves the aggregated reactivity score, $\eta<0.5$ reduces residual activation, and $\eta>0.5$ increases it.
When no pedestrian is spatially relevant, $\sum_i w_{i,t}=0$ and Eq.~\eqref{eq:adaptive_scale} gives $\kappa_t=1$, so the robot defaults to efficient free-space behavior.


\paragraph{Training and Deployment.}\label{sec:training_deployment}
The framework is trained in three stages.
First, we train the conservative backbone policy on non-reactive pedestrians and freeze its weights.
Second, with the backbone frozen, we train the low-rank residual modules on reactive-crowd rollouts to learn cooperative behavior without overwriting the conservative prior.
Third, we train the reactivity estimator using simulator labels and refine it with DAgger to reduce closed-loop covariate shift.
The backbone and residual parameters remain fixed throughout this stage.
DAgger refinement improves average closed-loop reactivity-estimation accuracy from 77.27\% to 84.78\%.
At deployment, no online gradient update is performed.
The robot estimates local human reactivity, computes $\kappa_t$, evaluates the reactivity-conditioned policy in Eq.~\eqref{eq:policy_family}, and executes the velocity command.
Algorithm~\ref{alg:reactivity_conditioned_adaptation} in Appendix~\ref{sec:implementation_details} summarizes deployment.
Appendices~\ref{sec:implementation_details} and \ref{sec:predictor_ablation_revised} provide implementation and estimator details.

\section{Evaluation under Human--Robot Interaction Shift}\label{sec:experiments_revised}
We evaluate whether \ours adjusts navigation to shifts in pedestrian response dynamics and whether learned reactivity estimates provide a useful conditioning signal. 
Simulation tests the complete pipeline across five response regimes, recorded human trajectories test estimation, and controlled robot trials isolate residual adaptation. 
Together, these evaluations separate adaptation to interaction shift, estimation quality, and hardware validation of the modulation mechanism.

\paragraph{Experimental Setup and Baselines.}
We evaluate in CrowdNav~\citep{chen2019crowd} with 20 pedestrians in a $12\,\mathrm{m}\times12\,\mathrm{m}$ environment with randomized starts and goals.
The robot radius is $0.3\,\mathrm{m}$, pedestrian radii are sampled from $[0.3,0.5]\,\mathrm{m}$, and all agents have a maximum speed $1.0\,\mathrm{m/s}$.
We report Success Rate (SR), Collision Rate (CR), Navigation Time (NT), and Path Length (PL).
Metric definitions appear in Appendix~\ref{sec:metrics_appendix}, and Timeout Rate (TR) follows from $\mathrm{TR}=100\%-\mathrm{SR}-\mathrm{CR}$.
All pedestrians use ORCA~\citep{van2011reciprocal}: reactive pedestrians account for the robot, while non-reactive pedestrians respond only to other pedestrians.
We test five regimes: homogeneous \emph{Non-reactive} and \emph{Reactive} crowds, \emph{Mixed} crowds with equal proportions of randomly distributed reactive and non-reactive pedestrians, \emph{Spatial Clusters} with regionally separated types, and \emph{Time-varying} crowds initialized as Mixed with independent $5\%$ reactivity-switching probability per pedestrian per timestep.
Labels remain fixed within episodes in the first four regimes.
Baselines comprise ORCA~\citep{van2011reciprocal}, BRNE~\citep{muchen2025mixed}, CrowdNav++~\citep{liu2023intention}, and GenSafeNav~\citep{yao2025towards} variants.
CrowdNav++ and GenSafeNav (Cons.) are trained on non-reactive crowds, GenSafeNav (Naive) on reactive crowds, and GenSafeNav (Rand.) across all five regimes.
\ours freezes the Cons. backbone and learns LoRA residuals on reactive crowds, selecting $\kappa_t$ online from estimated reactivity.
The main comparisons use neutral adaptation ($\eta=0.5$).
Navigation results use $1{,}250$ episodes per scenario across five seeds, reported as mean $\pm$ standard deviation across seeds.

\paragraph{Reactivity-Conditioned Adaptation and the Safety--Efficiency Trade-off.}


\begin{figure}[!b]
    \centering
    \begin{minipage}[t]{0.46\textwidth}
        \vspace{0pt}
        \centering
        \includegraphics[width=\linewidth]
        {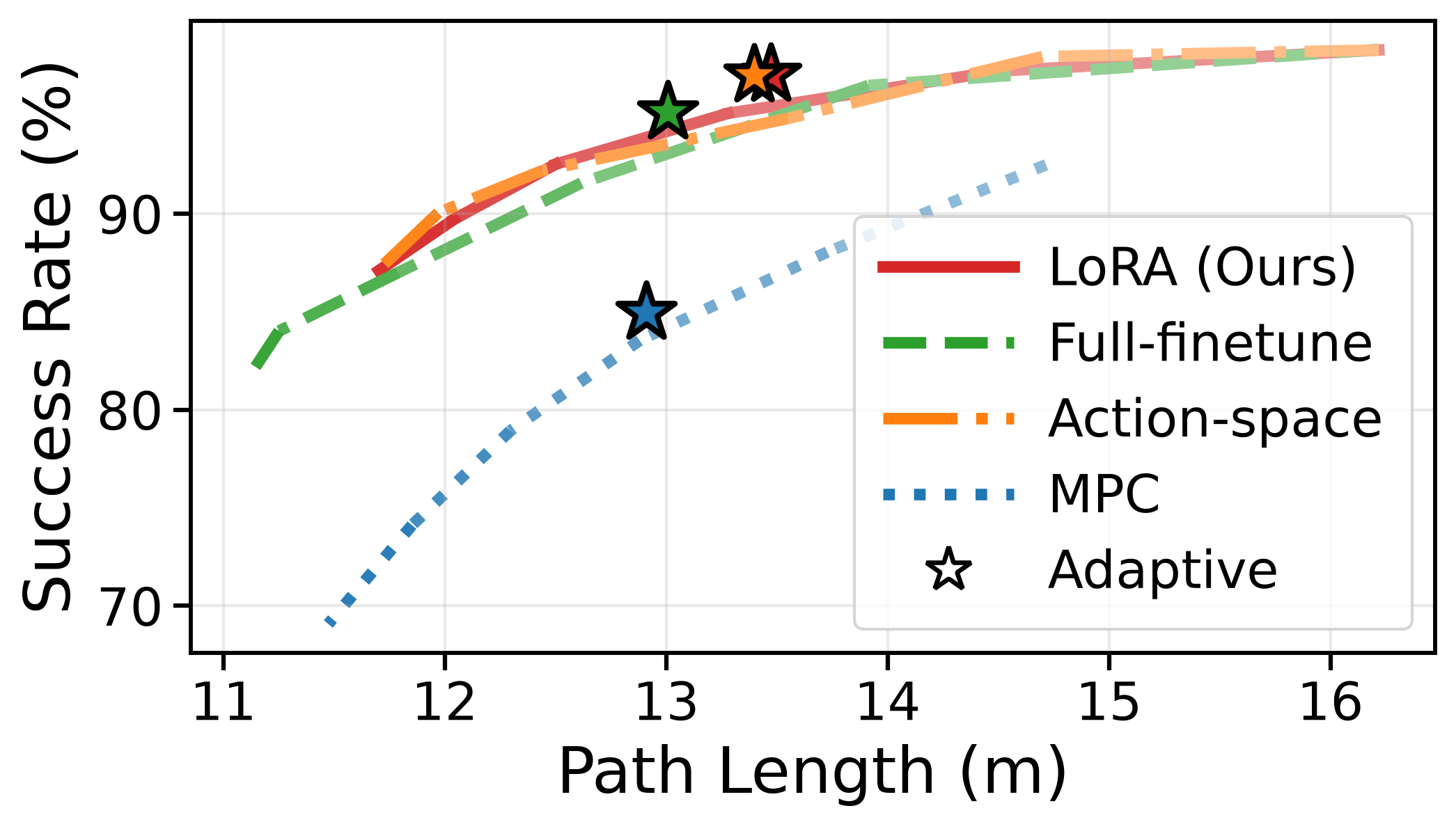}
        \caption{\textbf{Scenario-averaged adaptation trade-offs.}
        Fixed scales increase from light ($\kappa=0$) to dark ($\kappa=1$).
        Stars denote adaptation using ground-truth reactivity.}
        \label{fig:adaptation_tradeoff_appendix}
    \end{minipage}
    \hfill
    \begin{minipage}[t]{0.52\textwidth}
        \vspace{0pt}
        \centering
        \captionof{table}{\textbf{Reactivity estimation accuracy.}
        BDDS~\citep{hosseinzadeh2021toward} versus our learned estimator.
        FP and FN are percentages of all evaluated pedestrian--timestep pairs.}
        \label{tab:reactivity_accuracy}

        \small
        \setlength{\tabcolsep}{3pt}
        \resizebox{\linewidth}{!}{%
        \begin{tabular}{lcc}
            \toprule
            \textbf{Scenario} & \textbf{BDDS} & \textbf{Ours} \\
            & Acc./FP/FN (\%) & Acc./FP/FN (\%) \\
            \midrule
            Non-reactive
            & $72.6/27.4/0.0$ & $\mathbf{87.4}/12.6/0.0$ \\
            Reactive
            & $72.5/0.0/27.5$ & $\mathbf{90.4}/0.0/9.6$ \\
            Mixed
            & $72.8/12.1/15.1$ & $\mathbf{84.6}/7.0/8.4$ \\
            Spatial Clusters
            & $72.1/12.4/15.5$ & $\mathbf{84.4}/7.1/8.5$ \\
            Time-varying
            & $70.8/13.3/15.9$ & $\mathbf{77.1}/10.4/12.5$ \\
            \midrule
            Mean
            & $72.16/13.04/14.80$ & $\mathbf{84.78}/7.42/7.80$ \\
            \bottomrule
        \end{tabular}%
        }
    \end{minipage}
\end{figure}

Fig.~\ref{fig:adaptation_tradeoff_appendix} compares adaptation mechanisms using ground-truth reactivity.
Across LoRA, gated full fine-tuning, action-space interpolation, and MPC, increasing the fixed scale from $\kappa_t=0$ toward $\kappa_t=1$ generally shortens paths at the cost of lower SR, exposing the safety--efficiency trade-off associated with fixed adaptation strength.
Adaptive scaling offers a favorable safety--efficiency trade-off by conditioning policy adaptation or MPC clearance on local reactivity, retaining conservative behavior near non-reactive pedestrians while exploiting cooperative responses.
These results support reactivity conditioning across both learned policies and MPC.
Varying $\eta$ biases adaptation toward safety or efficiency, we evaluate this effect alongside the LoRA scaling sweep in Appendix~\ref{sec:full_lora_appendix_revised}.
Implementation details for all four adaptation mechanisms are provided in Appendix~\ref{sec:adaptation_mechanism_ablation}.

\begin{table*}[!t]
    \centering
\caption{\textbf{Effect of reactivity estimation on navigation.}
We compare BDDS's Bayesian estimator with our learned estimator,
keeping the backbone, LoRA residual, and spatial weighting fixed
and changing only the supplied reactivity estimates.}
    \label{tab:combined_performance}

    \resizebox{\textwidth}{!}
    {%
    \begin{tabular}{c | c | cccc}
    \toprule
    \textbf{Scenario}
    & \textbf{Conditioning Strategy}
    & \textbf{SR (\%)} $\uparrow$
    & \textbf{CR (\%)} $\downarrow$
    & \textbf{NT (s)} $\downarrow$
    & \textbf{PL (m)} $\downarrow$ \\
    \midrule

\multirow{2}{*}{\shortstack{Non-reactive}}

    & BDDS \citep{hosseinzadeh2021toward} (Bayesian)
    & 93.60 $\pm$ 0.63
    & 6.40 $\pm$ 0.63
    & \textbf{15.08} $\pm$ 0.34
    & \textbf{15.02} $\pm$ 0.33 \\

    & Ours (Learned)
    & \textbf{95.52} $\pm$ 0.69
    & \textbf{4.48} $\pm$ 0.69
    & 15.86 $\pm$ 0.45
    & 15.80 $\pm$ 0.43 \\

\midrule

\multirow{2}{*}{Reactive}

    & BDDS \citep{hosseinzadeh2021toward} (Bayesian)
    & \textbf{99.68} $\pm$ 0.33
    & \textbf{0.32} $\pm$ 0.33
    & 11.84 $\pm$ 0.18
    & 11.80 $\pm$ 0.19 \\

    & Ours (Learned)
    & 99.28 $\pm$ 0.33
    & 0.72 $\pm$ 0.33
    & \textbf{11.02} $\pm$ 0.12
    & \textbf{11.01} $\pm$ 0.12 \\

\midrule

\multirow{2}{*}{Mixed}

    & BDDS \citep{hosseinzadeh2021toward} (Bayesian)
    & 93.60 $\pm$ 1.44
    & 6.40 $\pm$ 1.44
    & 14.10 $\pm$ 0.46
    & 14.07 $\pm$ 0.39 \\

    & Ours (Learned)
    & \textbf{96.08} $\pm$ 1.58
    & \textbf{3.92} $\pm$ 1.58
    & \textbf{13.68} $\pm$ 0.43
    & \textbf{13.64} $\pm$ 0.37 \\

\midrule

\multirow{2}{*}{\shortstack{Spatial\\Clusters}}

    & BDDS \citep{hosseinzadeh2021toward} (Bayesian)
    & 93.28 $\pm$ 1.04
    & 6.72 $\pm$ 1.04
    & 13.60 $\pm$ 0.38
    & 13.56 $\pm$ 0.41 \\

    & Ours (Learned)
    & \textbf{94.88} $\pm$ 1.00
    & \textbf{5.12} $\pm$ 1.00
    & \textbf{13.01} $\pm$ 0.33
    & \textbf{12.98} $\pm$ 0.36 \\

\midrule

\multirow{2}{*}{\shortstack{Time-\\varying}}

    & BDDS \citep{hosseinzadeh2021toward} (Bayesian)
    & 93.44 $\pm$ 1.59
    & 6.56 $\pm$ 1.59
    & 14.72 $\pm$ 0.44
    & 14.66 $\pm$ 0.43 \\

    & Ours (Learned)
    & \textbf{94.48} $\pm$ 1.63
    & \textbf{5.52} $\pm$ 1.63
    & \textbf{14.19} $\pm$ 0.36
    & \textbf{14.13} $\pm$ 0.39 \\

    \bottomrule
    \end{tabular}%
    }\vspace{0.15cm}

    \raggedright \small
    \textit{*Note: Bold denotes the better mean among the two conditioning strategies.}
\end{table*}




\paragraph{Reactivity Belief Estimation under Interaction Shift.}
\label{sec:estimation_main}

Our learned estimator improves reactivity identification under interaction shift, providing a more accurate signal for online adaptation.
We evaluate it against BDDS's passive inference component~\citep{hosseinzadeh2021toward}, which assumes known pedestrian goals, an assumption often unrealistic in deployment.
BDDS interprets observed actions through a fixed goal-and-safety cost model, whereas our estimator learns temporal motion patterns conditioned on robot context and interactions among nearby pedestrians.
This context can help distinguish motion changes caused by other pedestrians from robot-directed responses.
Table~\ref{tab:reactivity_accuracy} reports scenario-averaged accuracy of \textbf{84.78\%} for our estimator versus $72.16\%$ for BDDS.
In Non-reactive crowds, the false-positive share decreases from $27.4\%$ to $12.6\%$.
In Reactive crowds, the false-negative share decreases from $27.5\%$ to $9.6\%$.

These improved estimates also benefit navigation, reducing collisions while shortening paths relative to BDDS conditioning in heterogeneous and time-varying crowds.
The two error types have different navigation implications: false positives can encourage reliance on cooperation that does not occur, while false negatives can cause unnecessary conservatism.
To isolate the effect of reactivity estimation, we change only the reactivity signal while keeping the backbone, residual, and spatial weighting fixed.
As shown in Table~\ref{tab:combined_performance}, averaged equally across Mixed, Spatial Clusters, and Time-varying, \ours with learned conditioning reduces CR from $6.56\%$ to $4.85\%$ and PL from $14.10$ to $13.58$~m compared with BDDS conditioning, relative reductions of $26.0\%$ and $3.6\%$, respectively.
Across all five scenarios, its mean SR approaches ground-truth conditioning ($96.05\%$ versus $97.10\%$), while scenario-averaged NT and PL each differ by less than $0.3\%$, as reported in Appendix~\ref{sec:full_lora_appendix_revised}.
On held-out PeRoI~\citep{agrawal2026peroi} moving-robot trajectories, using avoidance and neutrality annotations as proxies for reactive and non-reactive behavior, our estimator achieves $73.0\%$ balanced accuracy versus $68.8\%$ for tuned BDDS.
Baseline details and evaluation protocols appear in Appendix~\ref{sec:predictor_ablation_revised}.

\paragraph{Comparison with Navigation Baselines.}
\begin{table*}[!t]
    \centering
    \caption{\textbf{Interaction-shift benchmark.}
    We compare all methods across non-reactive, reactive, mixed,
    spatial-cluster, and time-varying scenarios over $1{,}250$
    test episodes per scenario across five random seeds.
    \ours improves the empirical safety--efficiency trade-off.}
    \label{tab:ood_results}

    \resizebox{\textwidth}{!}
    {
    \begin{tabular}{c|c|cccc}
    \toprule
    \textbf{Scenario} & \textbf{Methods}
    & \textbf{SR} (\%) $\uparrow$
    & \textbf{CR} (\%) $\downarrow$
    & \textbf{NT} (s) $\downarrow$
    & \textbf{PL} (m) $\downarrow$ \\
    \midrule

\multirow{7}{*}{\shortstack{Non-reactive}}


 & \textit{ORCA} \citep{van2011reciprocal}
 & \textit{66.40 $\pm$ 3.88}
 & \textit{29.20 $\pm$ 2.71}
 & \textit{23.68 $\pm$ 0.82}
 & \textit{15.03 $\pm$ 0.35} \\

 & \textit{BRNE} \citep{muchen2025mixed}
 & \textit{23.04 $\pm$ 1.56}
 & \textit{76.00 $\pm$ 1.17}
 & \textit{17.07 $\pm$ 0.54}
 & \textit{16.58 $\pm$ 0.53} \\

 & \textit{CrowdNav++} \citep{liu2023intention}
 & \textit{79.20 $\pm$ 2.35}
 & \textit{20.80 $\pm$ 2.35}
 & \textit{14.42 $\pm$ 0.28}
 & \textit{13.53 $\pm$ 0.22} \\

 & GenSafeNav (Cons.) \citep{yao2025towards}
 & \textbf{97.20} $\pm$ 0.40
 & \textbf{2.80} $\pm$ 0.40
 & 16.59 $\pm$ 0.47
 & 16.52 $\pm$ 0.47 \\

 & \textit{GenSafeNav (Naive)} \citep{yao2025towards}
 & \textit{59.52 $\pm$ 2.47}
 & \textit{40.48 $\pm$ 2.47}
 & \textit{11.61 $\pm$ 0.20}
 & \textit{11.58 $\pm$ 0.21} \\


 & GenSafeNav (Rand.) \citep{yao2025towards}
 & 91.92 $\pm$ 1.45
 & 7.84 $\pm$ 1.28
 & \textbf{15.16} $\pm$ 0.35
 & \textbf{15.05} $\pm$ 0.34 \\




 & Ours (RCRA)
 & \underline{95.52} $\pm$ 0.69
 & \underline{4.48} $\pm$ 0.69
 & \underline{15.86} $\pm$ 0.45
 & \underline{15.80} $\pm$ 0.43 \\

\midrule

\multirow{7}{*}{Reactive}


 & ORCA \citep{van2011reciprocal}
 & 93.12 $\pm$ 1.11
 & 0.72 $\pm$ 0.18
 & 23.32 $\pm$ 0.72
 & 15.39 $\pm$ 0.72 \\

 & BRNE \citep{muchen2025mixed}
 & 86.80 $\pm$ 1.67
 & 2.40 $\pm$ 0.57
 & 18.26 $\pm$ 0.42
 & 17.75 $\pm$ 0.41 \\

 & CrowdNav++ \citep{liu2023intention}
 & 94.24 $\pm$ 0.46
 & 5.44 $\pm$ 0.46
 & 12.57 $\pm$ 0.17
 & 11.83 $\pm$ 0.16 \\

 & GenSafeNav (Cons.) \citep{yao2025towards}
 & \textbf{100.00} $\pm$ 0.00
 & \textbf{0.00} $\pm$ 0.00
 & 16.24 $\pm$ 0.47
 & 16.19 $\pm$ 0.47 \\

 & GenSafeNav (Naive) \citep{yao2025towards}
 & 99.20 $\pm$ 0.40
 & 0.80 $\pm$ 0.40
 & \textbf{10.69} $\pm$ 0.07
 & \textbf{10.66} $\pm$ 0.08 \\


 & GenSafeNav (Rand.) \citep{yao2025towards}
 & \underline{99.52} $\pm$ 0.52
 & \underline{0.16} $\pm$ 0.36
 & 14.01 $\pm$ 0.24
 & 13.92 $\pm$ 0.24 \\




 & Ours (RCRA)
 & 99.28 $\pm$ 0.33
 & 0.72 $\pm$ 0.33
 & \underline{11.02} $\pm$ 0.12
 & \underline{11.01} $\pm$ 0.12 \\

\midrule

\multirow{7}{*}{Mixed}


 & ORCA \citep{van2011reciprocal}
 & 83.68 $\pm$ 1.91
 & 11.84 $\pm$ 1.49
 & 23.87 $\pm$ 0.41
 & 15.46 $\pm$ 0.39 \\

 & \textit{BRNE} \citep{muchen2025mixed}
 & \textit{43.36 $\pm$ 5.20}
 & \textit{53.92 $\pm$ 4.93}
 & \textit{16.52 $\pm$ 0.78}
 & \textit{16.05 $\pm$ 0.76} \\

 & CrowdNav++ \citep{liu2023intention}
 & 85.44 $\pm$ 1.25
 & 14.56 $\pm$ 1.25
 & 13.71 $\pm$ 0.21
 & \underline{13.59} $\pm$ 0.14 \\

 & GenSafeNav (Cons.) \citep{yao2025towards}
 & \textbf{98.56} $\pm$ 0.73
 & \textbf{1.36} $\pm$ 0.73
 & 16.61 $\pm$ 0.58
 & 16.55 $\pm$ 0.57 \\

 & GenSafeNav (Naive) \citep{yao2025towards}
 & 82.40 $\pm$ 4.79
 & 17.60 $\pm$ 4.79
 & \textbf{11.36} $\pm$ 0.20
 & \textbf{11.34} $\pm$ 0.17 \\


 & GenSafeNav (Rand.) \citep{yao2025towards}
 & \underline{96.72} $\pm$ 0.66
 & \underline{3.20} $\pm$ 0.75
 & 14.74 $\pm$ 0.35
 & 14.64 $\pm$ 0.35 \\




 & Ours (RCRA)
 & 96.08 $\pm$ 1.58
 & 3.92 $\pm$ 1.58
 & \underline{13.68} $\pm$ 0.43
 & 13.64 $\pm$ 0.37 \\

\midrule

\multirow{7}{*}{\shortstack{Spatial\\Clusters}}


 & ORCA \citep{van2011reciprocal}
 & 89.60 $\pm$ 2.24
 & 6.88 $\pm$ 1.88
 & 18.94 $\pm$ 0.51
 & 14.14 $\pm$ 0.66 \\

 & \textit{BRNE} \citep{muchen2025mixed}
 & \textit{65.12 $\pm$ 4.27}
 & \textit{30.80 $\pm$ 3.97}
 & \textit{17.56 $\pm$ 0.42}
 & \textit{17.03 $\pm$ 0.41} \\

 & CrowdNav++ \citep{liu2023intention}
 & 85.60 $\pm$ 1.57
 & 14.40 $\pm$ 1.57
 & 13.31 $\pm$ 0.12
 & 13.13 $\pm$ 0.29 \\

 & GenSafeNav (Cons.) \citep{yao2025towards}
 & \underline{97.12} $\pm$ 0.77
 & \underline{2.88} $\pm$ 0.77
 & 15.47 $\pm$ 0.68
 & 15.40 $\pm$ 0.67 \\

 & GenSafeNav (Naive) \citep{yao2025towards}
 & 82.88 $\pm$ 2.42
 & 17.12 $\pm$ 2.42
 & \textbf{10.83} $\pm$ 0.16
 & \textbf{10.80} $\pm$ 0.16 \\


 & GenSafeNav (Rand.) \citep{yao2025towards}
 & \textbf{97.60} $\pm$ 0.28
 & \textbf{2.40} $\pm$ 0.28
 & 14.79 $\pm$ 0.34
 & 14.65 $\pm$ 0.35 \\




 & Ours (RCRA)
 & 94.88 $\pm$ 1.00
 & 5.12 $\pm$ 1.00
 & \underline{13.01} $\pm$ 0.33
 & \underline{12.98} $\pm$ 0.36 \\

\midrule

\multirow{7}{*}{\shortstack{Time-\\varying}}


 & ORCA \citep{van2011reciprocal}
 & 82.56 $\pm$ 2.41
 & 12.40 $\pm$ 1.41
 & 23.66 $\pm$ 1.04
 & 15.11 $\pm$ 0.53 \\

 & \textit{BRNE} \citep{muchen2025mixed}
 & \textit{47.84 $\pm$ 2.81}
 & \textit{49.68 $\pm$ 3.18}
 & \textit{17.50 $\pm$ 0.79}
 & \textit{17.01 $\pm$ 0.77} \\

 & CrowdNav++ \citep{liu2023intention}
 & 85.28 $\pm$ 2.07
 & 14.72 $\pm$ 2.07
 & 14.28 $\pm$ 0.18
 & \underline{14.04} $\pm$ 0.17 \\

 & GenSafeNav (Cons.) \citep{yao2025towards}
 & \textbf{98.88} $\pm$ 0.77
 & \textbf{1.12} $\pm$ 0.77
 & 16.50 $\pm$ 0.53
 & 16.44 $\pm$ 0.53 \\

 & GenSafeNav (Naive) \citep{yao2025towards}
 & 81.84 $\pm$ 4.20
 & 18.16 $\pm$ 4.20
 & \textbf{11.60} $\pm$ 0.22
 & \textbf{11.55} $\pm$ 0.21 \\


 & GenSafeNav (Rand.) \citep{yao2025towards}
 & \underline{96.96} $\pm$ 1.08
 & \underline{2.72} $\pm$ 1.00
 & 14.74 $\pm$ 0.29
 & 14.65 $\pm$ 0.29 \\




 & Ours (RCRA)
 & 94.48 $\pm$ 1.63
 & 5.52 $\pm$ 1.63
 & \underline{14.19} $\pm$ 0.36
 & 14.13 $\pm$ 0.39 \\

    \bottomrule
    \end{tabular}
    }\vspace{0.15cm}

    \raggedright \small
    \textit{*Note: Performance rankings
    (\textbf{1st} and \underline{2nd}) are calculated among methods
    that achieve SR $>80\%$.}
\end{table*}

\ours combines efficient navigation where pedestrians cooperate with conservative behavior where they do not, balancing cooperation and caution as local interaction conditions change.
Table~\ref{tab:ood_results} compares \ours with the navigation baselines.
In Reactive crowds, its PL approaches GenSafeNav (Naive)'s ($11.01$~m versus $10.66$~m), with comparable SR ($99.28\%$ versus $99.20\%$).
In Non-reactive crowds, its SR remains close to GenSafeNav (Cons.) ($95.52\%$ versus $97.20\%$), whereas GenSafeNav (Naive) drops to $59.52\%$.
\ours balances efficiency and success when reactive and non-reactive pedestrians coexist or their responses change over time.
Averaged across Mixed, Spatial Clusters, and Time-varying, it shortens paths by $15.8\%$ relative to GenSafeNav (Cons.), with mean SR of $95.15\%$ compared with $98.19\%$ for GenSafeNav (Cons.) and $82.37\%$ for GenSafeNav (Naive).
Across all five scenarios, \ours improves overall navigation efficiency while maintaining a high mean SR.
Compared with GenSafeNav (Cons.) and (Rand.), \ours reduces scenario-averaged NT by \textbf{16.8\%} and \textbf{7.7\%}, and PL by \textbf{16.7\%} and \textbf{7.3\%}, respectively, with mean SR of $96.05\%$ versus $98.35\%$ and $96.54\%$.

Explicit local reactivity conditioning also offers benefits beyond training across varied response regimes alone.
GenSafeNav (Rand.), trained across all five scenarios without explicit reactivity estimation, achieves lower SR in Non-reactive crowds ($91.92\%$ versus $95.52\%$ for \ours) and longer paths in Reactive crowds ($13.92$~m versus $11.01$~m), despite comparable SR.
These results suggest that varied training conditions alone do not fully capture the benefits of adapting to locally relevant pedestrian responses.
ORCA and BRNE achieve lower SR in Non-reactive crowds, consistent with their interaction models anticipating collision-avoidance responses from pedestrians.
CrowdNav++ lacks an explicit safety-cost constraint and reaches a worst-case SR of $79.20\%$.

\paragraph{Controlled Real-Robot Validation } \label{sec:real_robot}

We deploy the navigation stack on a Clearpath Jackal running ROS 2 with an onboard NVIDIA RTX 4000 GPU.
Pedestrians are detected from 10~Hz Velodyne LiDAR scans using DR-SPAAM~\citep{jia2020dr}, tracked using SORT~\citep{bewley2016simple}, and their trajectories are predicted using GST~\citep{huang2021learning}.
A command shim maps the policy’s holonomic $(v_x,v_y)$ outputs to the Jackal's unicycle commands $(v,\omega)$.
Both virtual trials and physical demonstrations use three pedestrians: one non-reactive and two reactive.
ORCA-controlled virtual pedestrians provide 30 quantitative trials per policy.
Physical demonstrations use pedestrians instructed to maintain nominal motion (non-reactive) or yield to the robot (reactive).
We compare GenSafeNav (Cons.), GenSafeNav (Naive), and \ours with a target 5~m ahead of the robot's initial position.
Prescribed reactivity labels isolate policy adaptation from estimation errors.
\begin{figure}[!t]
 \centering
 \includegraphics[width=\linewidth]{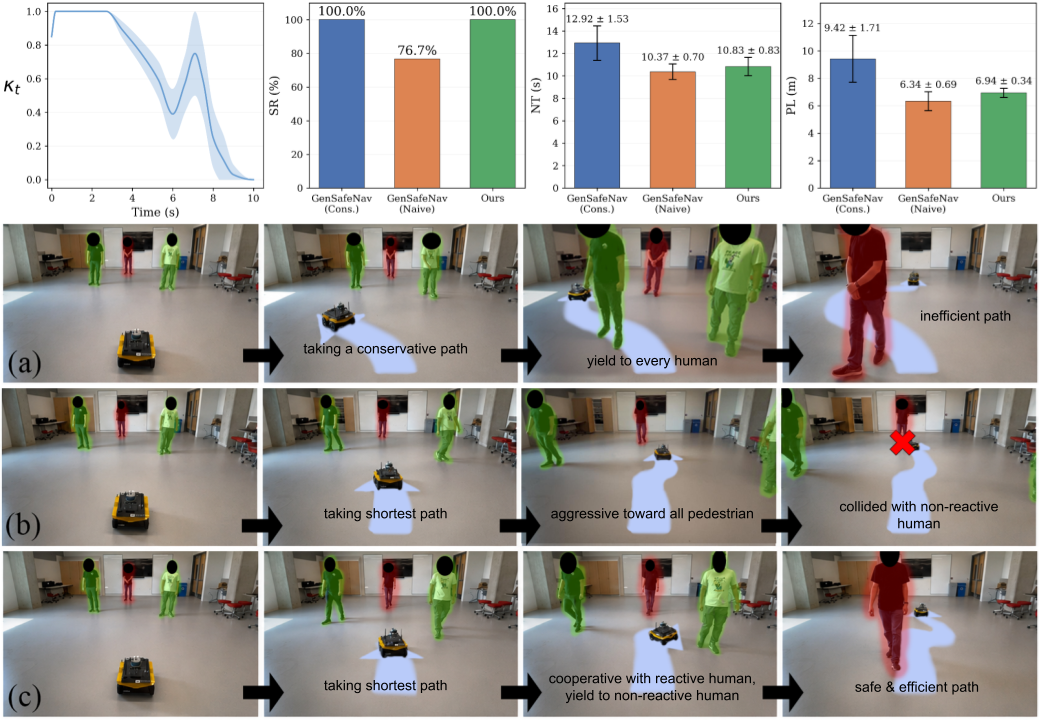}
\caption{\textbf{Controlled hardware validation with prescribed reactivity labels.} 
Top: representative adaptation scale $\kappa_t$ and quantitative performance over 30 virtual-pedestrian trials per policy.
Bottom: representative demonstrations with instructed physical pedestrians using (a) GenSafeNav (Cons.), (b) GenSafeNav (Naive), and (c) \ours.}
 \label{fig:hardware_exp}
\end{figure}

Fig.~\ref{fig:hardware_exp} illustrates a conservative detour with GenSafeNav (Cons.) and a direct route ending in collision with GenSafeNav (Naive).
\ours yields to the non-reactive pedestrian while exploiting gaps created by reactive pedestrians.
As the non-reactive pedestrian becomes locally relevant, $\kappa_t$ decreases from approximately one toward zero, shifting behavior toward the conservative backbone.
Across 30 virtual-pedestrian trials per policy, \ours achieves 30 successful runs versus 23 for GenSafeNav (Naive), reducing NT by 16.2\% and PL by 26.3\% relative to GenSafeNav (Cons.).
These experiments validate adaptation using prescribed reactivity labels.
Closed-loop deployment with learned reactivity estimation remains future work, potentially incorporating body orientation and gaze as complementary cues to robot-directed attention.

\section{Discussion and Limitations}\label{sec:conclusion_revised}

Our results support explicit pedestrian-reactivity estimation as a useful signal for modulating a pretrained navigation policy under heterogeneous and changing local interactions.
\ours spatially aggregates these estimates to control a learned residual around a frozen conservative backbone.
With navigation components held fixed, learned conditioning reduces collision rate and PL relative to BDDS conditioning in heterogeneous and time-varying crowds.
\noindent\textbf{Limitations.}
The reactivity estimator relies on binary supervision and motion history and may miss ambiguous real-world reactivity. It is sensitive to measurement noise and differences between simulated and real-robot motion.
Future work will evaluate the learned estimator in closed-loop deployment and incorporate body orientation and gaze.

\bibliography{example}

\clearpage

\appendix

\section{Additional Related Work}\label{sec:related_appendix}

\paragraph{Robot Crowd Navigation.}
Classical methods such as Social Force, velocity obstacles, and reciprocal collision avoidance generate local motion using structured assumptions about pedestrian behavior~\citep{helbing1995social,fiorini1998motion,van2011reciprocal}, while cooperative planners explicitly account for human responses to robot motion~\citep{trautman2010unfreezing,trautman2015robot}.
Learning-based methods capture crowd interactions through multi-agent reinforcement learning, attention-based pedestrian representations, and spatio-temporal interaction models~\citep{chen2017socially,everett2018motion,chen2019crowd,liu2021decentralized}.
Trajectory prediction further enables policies to anticipate pedestrian motion~\citep{liu2023intention}, while constrained reinforcement learning incorporates safety costs and prediction uncertainty~\citep{yao2025towards}.
However, the resulting policies can depend on the human--robot interaction patterns encountered during training: policies are often trained under the assumption that pedestrians do not react to the robot, encouraging safe but conservative behavior.
Recent approaches such as Human--Human \& Human--Robot Interaction Transformer \citep{shen2026human} train across varied pedestrian-response conditions, but do not explicitly adapt the policy online based on estimated pedestrian reactivity.
Model-based approaches such as Behavior-Driven Danger Signaling (BDDS)~\citep{hosseinzadeh2021toward} address uncertainty about human responses through Bayesian inference, but require a human goal to evaluate their behavioral cost model. 
Such goals may be unknown or change during real-world interactions, limiting the applicability of this assumption. 
We instead estimate robot-directed reactivity from observed interaction history and use these estimates to modulate a learned residual on a frozen conservative policy as interaction conditions change.

\paragraph{Prediction uncertainty and conformal safety.}
Pedestrian trajectory prediction is central to social navigation, with representative models including Social-LSTM, Social-GAN, and Trajectron++~\citep{alahi2016social,gupta2018social,salzmann2020trajectron}.
Broader surveys further categorize human-motion prediction methods by interaction modeling, uncertainty handling, and dataset assumptions, highlighting why prediction remains difficult under distribution shift~\citep{rudenko2020human}.
Conformal prediction provides distribution-free coverage under assumptions such as exchangeability, standard guarantees do not automatically hold under arbitrary distribution shift~\citep{angelopoulos2021gentle}.
Recent crowd-navigation methods use adaptive or interaction-sensitive conformal uncertainty to regulate planning under out-of-distribution pedestrian behavior~\citep{yao2025towards,huang2025interaction}.
These methods regulate navigation using prediction uncertainty. Our approach additionally estimates robot-directed reactivity, because a deviation caused by yielding may create usable free space and warrant a different navigation response.

\paragraph{Policy adaptation and parameter-efficient tuning.}
Robotic policies can be robustified or adapted through domain randomization, meta-learning, online reinforcement learning, imitation-learning correction, and efficient policy modulation~\citep{tobin2017domain,peng2018sim,finn2017model,ross2011reduction,xu2023hyper,schmied2023learning}.
However, adapting the entire policy can require longer training and greater computational resources. 
To reduce this overhead, parameter-efficient tuning freezes most pretrained parameters and trains a small set of additional parameters, as in residual adapters, neural-network adapters, low-rank adaptation (LoRA), and adapters for navigation and robot learning~\citep{rebuffi2017learning,houlsby2019parameter,hu2022lora,qiao2023vln,liu2024tail,team2024octo}.

\section{Evaluation Metrics}
\label{sec:metrics_appendix}

Each episode terminates in exactly one of three outcomes: success, collision, or timeout. Let $N_{\mathrm{total}}$, $N_{\mathrm{success}}$, and $N_{\mathrm{collision}}$ denote the corresponding total and outcome counts. We report:
\begin{itemize}
 \item \textbf{Success Rate (SR):} percentage of episodes that reach the goal before collision or timeout, $\mathrm{SR}=100\%\,N_{\mathrm{success}}/N_{\mathrm{total}}$.
 \item \textbf{Collision Rate (CR):} percentage of episodes terminated by collision, $\mathrm{CR}=100\%\,N_{\mathrm{collision}}/N_{\mathrm{total}}$. The timeout percentage is $\mathrm{TR}=100\%-\mathrm{SR}-\mathrm{CR}$.
 \item \textbf{Navigation Time (NT):} mean duration in seconds over successful episodes only, $\mathrm{NT}=N_{\mathrm{success}}^{-1}\sum_{k=1}^{N_{\mathrm{success}}}T_k$, where $T_k$ is the completion time of successful episode $k$.
 \item \textbf{Path Length (PL):} mean traveled distance in meters over successful episodes only, $\mathrm{PL}=N_{\mathrm{success}}^{-1}\sum_{k=1}^{N_{\mathrm{success}}}L_k$, where $L_k$ is the total distance traveled by the robot until reaching the goal in successful episode $k$.
\end{itemize}
We compute scenario-averaged metrics as unweighted means across the five scenarios, percentage reductions compare these means. 
NT and PL are computed over successful episodes, whose composition may differ across methods, and should therefore be interpreted alongside SR and CR.

\section{Implementation Details}
\label{sec:implementation_details}

\paragraph{Base architecture and simulator.}
We build on the GenSafeNav~\citep{yao2025towards} crowd-navigation framework, which uses GST trajectory prediction, prediction-uncertainty features, human--human attention, human--robot attention, temporal fusion, and actor--critic layers.
We retain GenSafeNav's reward and safety-cost formulations and PPO-Lagrangian training settings, with a cost limit of 0.4.
We use a prediction horizon of $K=5$ timesteps, a spatial weighting radius of $R=5\,\mathrm{m}$, and $\epsilon=10^{-6}$ in the adaptation scale.

\begin{algorithm}[!t]
\caption{Reactivity-conditioned residual adaptation}
\label{alg:reactivity_conditioned_adaptation}
\begin{algorithmic}[1]
\STATE \textbf{Input:} frozen parameters $\theta_0$, residual $\Delta\theta$, estimator $f_\phi$, prediction horizon $K$, history length $H$, sensing radius $R$, bias-control parameter $\eta\in(0,1)$, and $\epsilon>0$
\FOR{each control step $t$}
 \STATE Construct $o_t$ from robot state, tracked pedestrians, predictions, and uncertainty features
 \FOR{each observed pedestrian $i=1,\ldots,N_t$}
 \STATE Estimate reactivity belief $\hat{b}_{i,t}$ using Eq. \eqref{eq:estimate}
 \STATE Compute spatial relevance $w_{i,t}$ using Eq.~\eqref{eq:spatial_weight_revised}
 \ENDFOR
 \STATE Aggregate continuous beliefs into $\kappa_t$ using Eq.~\eqref{eq:adaptive_scale}
 \STATE Set adapted weights $W_{\ell,t}=W_{\ell,0}+\kappa_t\Delta W_{\ell}$ for all $\ell\in\mathcal{L}_{\mathrm{adapt}}$
 \STATE Sample action $a_t\sim\pi_{\kappa_t}(a_t\mid o_t)$
 \STATE Execute $a_t$
\ENDFOR
\end{algorithmic}
\end{algorithm}

\paragraph{Three-stage training pipeline.}
Our framework is trained in three stages. First, we initialize the conservative backbone from a pretrained GenSafeNav checkpoint trained for $2\times10^7$ environment steps and freeze all backbone parameters. Second, we fine-tune only the LoRA residual modules on reactive-crowd rollouts for $2\times10^7$ environment steps, using the same PPO-Lagrangian hyperparameters as GenSafeNav. During this stage, the residual branch is fully activated with $\kappa_t=1$. Third, we train the reactivity estimator using simulator-provided reactivity labels and refine it with DAgger. All simulation experiments are run on an NVIDIA RTX PRO 6000 Blackwell GPU.

\paragraph{LoRA adaptation.}
LoRA modules are inserted into the human--robot cross-attention feature projections and the linear layers of the actor and critic MLPs, while the remaining GenSafeNav backbone remains frozen. The LoRA adapters use rank $r=1$ and $\alpha=128$.
The fixed gain $\alpha/r$ is absorbed into the factors in our weight-update notation. The backbone contains approximately 2.5M parameters, while LoRA introduces approximately 3.5K trainable parameters, equivalent to $0.14\%$ of the backbone.
At deployment, the effective adapted weights are controlled by the online scale $\kappa_t$, allowing the policy to interpolate between the frozen conservative backbone and the learned reactive-crowd residual.

\section{Reactivity Estimation: Baseline and Evaluation Details}
\label{sec:predictor_ablation_revised}

Our reactivity estimator is inspired by the Human Scene Transformer (HST)~\citep{salzmann2023robots}, but adapts its architecture to reactivity classification rather than trajectory forecasting.
We remove the visual cues used by HST and replace the original scene context with the robot state.
The estimator takes a short history of human motion together with the robot state.
Feature-level attention first encodes human motion features, followed by per-agent temporal alignment and joint self-attention over agent--time tokens to capture human--human and human--robot interactions.
We use four attention heads, six latent modes, dropout $0.1$, and a learned temporal embedding.
The hidden dimension is 128 in simulation and 32 for PeRoI, with a feed-forward dimension of 128 in both settings.
The representation at each pedestrian's last valid observation is used for classification.
Latent-mode representations are averaged before a LayerNorm--MLP classification head produces one logit per pedestrian.
A sigmoid maps this logit to the continuous reactivity belief $\hat b_{i,t}\in[0,1]$.
The estimator uses an approximately 2\,s observation history in both domains: $H=8$ samples in simulation at $\Delta t=0.25$\,s and $H=30$ samples for PeRoI at $\Delta t\approx0.0667$\,s.
We train for 30 epochs with class-weighted binary cross-entropy, AdamW with learning rate $3\times10^{-4}$ and weight decay $10^{-4}$, a cosine learning-rate schedule, batch size 32, and gradient clipping at $1.0$.

\paragraph{Evaluation metrics.}
For classification, both estimators predict reactive when their continuous score exceeds $0.5$ and non-reactive otherwise.
True positives (TP) and true negatives (TN) denote correctly classified reactive and non-reactive samples, respectively.
False positives (FP) are non-reactive samples incorrectly classified as reactive, while false negatives (FN) are reactive samples incorrectly classified as non-reactive.
Simulation accuracy is the percentage of correctly classified pedestrian--timestep pairs, averaged equally across the five scenarios.
For class-imbalanced PeRoI evaluation and BDDS parameter selection, we use balanced accuracy, $50\%\left[\mathrm{TP}/(\mathrm{TP}+\mathrm{FN}) +\mathrm{TN}/(\mathrm{TN}+\mathrm{FP})\right]$.
The classification threshold is used only for reporting and does not enter the continuous navigation scale in Eq.~\eqref{eq:adaptive_scale}.

\paragraph{Behavior-Driven Danger Signaling (BDDS).}
We adapt the Bayesian estimator of \citet{hosseinzadeh2021toward}, which updates a belief over whether pedestrian actions minimize a goal cost alone or a combined goal and robot-safety cost.
We evaluate its passive inference component without emitting danger signals.
We denote its posterior score by $\hat b_{i,t}^{\mathrm{BDDS}}$, using our belief notation. This is the probability of the model's robot-safety-responsive hypothesis given the observed human--robot motion history, and serves as a proxy for our reactivity label.
We extend the paper's one-dimensional example to planar kinematics,
$p^h_{i,t+1}=p^h_{i,t}+u^h_{i,t}\Delta t$, where $u^h_{i,t}$ is a candidate planar velocity and $\Delta t$ is the sampling interval.
The action grid contains a stop action and speeds
$\{0.5,1,1.5,2\}v_H$ at seven headings from $-90^\circ$ to $90^\circ$
in $30^\circ$ increments relative to the current pedestrian heading,
giving 29 actions.
Here, $v_H$ is the reference pedestrian speed.
Observed velocity is assigned to the nearest action, and costs are evaluated at the candidate next position.
We use $(v_H,\Delta t)=(1.0\,\mathrm{m/s},0.25\,\mathrm{s})$ in simulation and
$(1.41\,\mathrm{m/s},0.0667\,\mathrm{s})$ on PeRoI, retaining the published cost weights and random-action mixture weight $\omega_H=0.1$.
We select the Boltzmann sharpness $\gamma$ by maximizing balanced accuracy on the training split and fix it for held-out evaluation:
$\gamma^*_{\mathrm{sim}}=1.44$ and
$\gamma^*_{\mathrm{PeRoI}}=0.30$.

\begin{figure}[!t]
    \centering
    \includegraphics[width=0.49\linewidth]{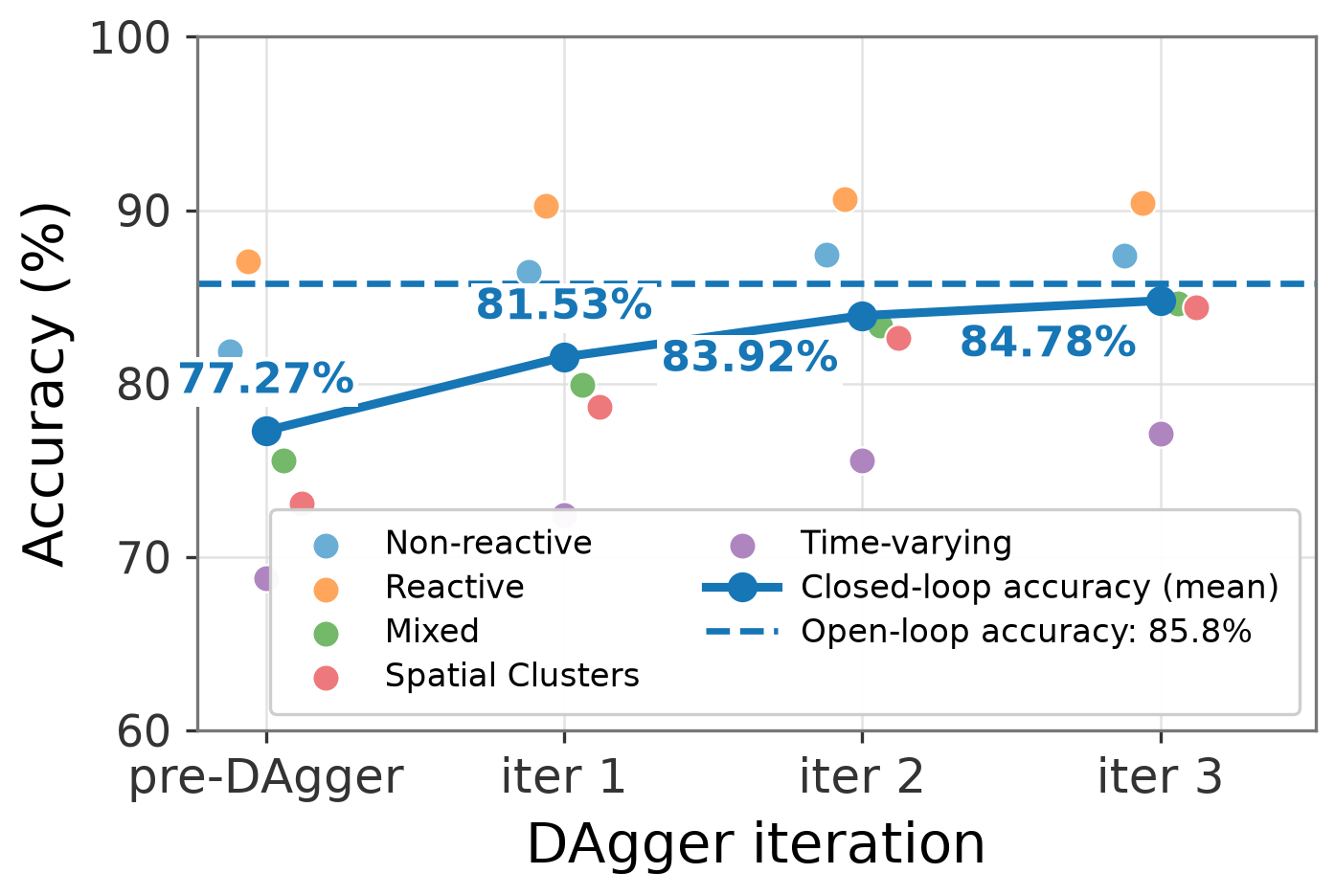}
    \caption{\textbf{DAgger refinement.}
    Reactivity estimation accuracy across aggregation iterations.}
    \label{fig:dagger_accuracy}
\end{figure}

\paragraph{Simulation data splits and DAgger protocol.}
Initial estimator training, DAgger data collection, and final evaluation use disjoint episode sets.
The offline accuracy reference is computed over $250$ evaluation episodes per scenario.
At each DAgger iteration, we collect $250$ episodes per scenario using the current adaptive policy, label the visited observations with simulator ground truth, and aggregate them with the existing training data to update the estimator.
Final estimator and navigation results are reported on held-out episodes excluded from training and DAgger collection.
Fig.~\ref{fig:dagger_accuracy} shows that the pre-DAgger estimator reaches only $77.27\%$ scenario-averaged closed-loop accuracy, indicating a substantial distribution mismatch.
After three DAgger iterations, accuracy improves to 84.78\%, a gain of 7.51 percentage points, approaching the offline evaluation reference of 85.80\%. Table~\ref{tab:reactivity_accuracy} reports the corresponding
scenario-wise accuracy and error breakdown.

\paragraph{Real-Human Trajectory Evaluation.}
\label{sec:peroi_estimation}

PeRoI (Pedestrian-Robot Interaction) contains recorded pedestrian trajectories from two outdoor sites, annotated as avoidance, neutrality, or attraction under absent, stationary, and moving-robot conditions~\citep{agrawal2026peroi}.
We use the moving-robot recording and map avoidance to reactive ($z_{i,t}=1$) and neutrality to non-reactive ($z_{i,t}=0$) throughout each annotated track. These track-level behavior labels provide a proxy for reactivity rather than timestep-specific causal labels.
Although attraction is a robot-directed response, approaching the robot differs from the yielding or avoidance behavior targeted by our adaptation signal.
We therefore exclude the 18 attraction tracks (approximately 8.0\% of 226 labeled tracks), leaving 208 tracks: 149 neutral and 59 avoidance.

We use a stratified 70/30 train--validation split by pedestrian track, keeping all windows from each track in the same split.
Velocities are derived from timestamps and positions, and tracking gaps split the sequences.
Because the dataset does not provide intended pedestrian goals, BDDS uses each contiguous track's final observed position as a goal proxy, giving it access to future endpoint information in this offline evaluation.
Both estimators are evaluated at the same window endpoints against the track-level annotations, using the $0.5$ classification threshold.
The validation split is excluded from estimator training and BDDS $\gamma$ selection.
Learned estimation is evaluated on PeRoI in Section~\ref{sec:estimation_main}.

\section{Adaptability and the Safety--Efficiency Trade-off}
\label{sec:full_lora_appendix_revised}

We examine how the policy adjusts its navigation behavior to robot-directed reactivity by comparing fixed LoRA scales and several adaptive mechanisms.
Table~\ref{tab:lora_ablation_full_revised} provides the fixed-scale LoRA sweep together with adaptive LoRA (Adaptive-Safe, Adaptive, and Adaptive-Efficient), gated full fine-tuning, action-space interpolation, and adaptive MPC.
A fixed global scale $\kappa$ remains constant across pedestrians, timesteps, and scenarios.
The adaptive variants use the runtime adaptation scale under ground-truth reactivity labels, isolating the adaptation mechanism from estimation errors.
Fig.~\ref{fig:adaptation_tradeoff_appendix} summarizes the scenario-averaged trade-offs: $\kappa=0$ uses the conservative backbone and $\kappa=1$ fully activates the residual, with intermediate scales tracing an empirically monotonic decrease in average path length and SR.

\begin{table*}[!b]
\centering
\caption{\textbf{Adaptation mechanism ablations.}
Fixed residual scales expose a safety--efficiency trade-off; adaptive LoRA, full fine-tuning, action-space interpolation, and MPC use ground-truth reactivity to isolate online conditioning. Adaptive-Safe and Adaptive-Efficient bias LoRA toward safety and efficiency, respectively, and are omitted in homogeneous scenarios.}
\label{tab:lora_ablation_full_revised}

\resizebox{\linewidth}{!}{%
\begin{tabular}{c|c|cccc}
\toprule
\textbf{Scenario}
& \textbf{Setting}
& \textbf{SR} (\%) $\uparrow$
& \textbf{CR} (\%) $\downarrow$
& \textbf{NT (s)} $\downarrow$
& \textbf{PL (m)} $\downarrow$ \\
\midrule

\multirow{7}{*}{Non-reactive}

& LoRA ($\kappa_t=0.2$)
& 93.84 $\pm$ 0.78
& 6.16 $\pm$ 0.78
& 15.01 $\pm$ 0.34
& 14.94 $\pm$ 0.34 \\

& LoRA ($\kappa_t=0.6$)
& 83.12 $\pm$ 1.78
& 16.88 $\pm$ 1.78
& 13.25 $\pm$ 0.25
& 13.23 $\pm$ 0.25 \\

& LoRA ($\kappa_t=1.0$)
& 73.12 $\pm$ 3.99
& 26.88 $\pm$ 3.99
& 12.35 $\pm$ 0.25
& 12.33 $\pm$ 0.25 \\

& LoRA (Adaptive)
& 97.04 $\pm$ 0.36
& 2.96 $\pm$ 0.36
& 16.08 $\pm$ 0.41
& 16.01 $\pm$ 0.44 \\

& Full fine-tuning (Adaptive)
& 97.04 $\pm$ 0.61
& 2.96 $\pm$ 0.61
& 16.01 $\pm$ 0.45
& 15.93 $\pm$ 0.45 \\

& Action-space (Adaptive)
& 97.04 $\pm$ 0.46
& 2.96 $\pm$ 0.46
& 16.00 $\pm$ 0.47
& 15.93 $\pm$ 0.46 \\

& MPC (Adaptive)
& 69.44 $\pm$ 2.31
& 30.56 $\pm$ 2.31
& 14.81 $\pm$ 0.38
& 14.22 $\pm$ 0.21 \\

\midrule

\multirow{7}{*}{Reactive}

& LoRA ($\kappa_t=0.2$)
& 99.92 $\pm$ 0.18
& 0.08 $\pm$ 0.18
& 14.34 $\pm$ 0.43
& 14.28 $\pm$ 0.42 \\

& LoRA ($\kappa_t=0.6$)
& 99.68 $\pm$ 0.33
& 0.32 $\pm$ 0.33
& 11.73 $\pm$ 0.27
& 11.71 $\pm$ 0.27 \\

& LoRA ($\kappa_t=1.0$)
& 99.20 $\pm$ 0.28
& 0.80 $\pm$ 0.28
& 10.87 $\pm$ 0.11
& 10.87 $\pm$ 0.11 \\

& LoRA (Adaptive)
& 99.20 $\pm$ 0.28
& 0.80 $\pm$ 0.28
& 10.87 $\pm$ 0.11
& 10.87 $\pm$ 0.11 \\

& Full fine-tuning (Adaptive)
& 99.20 $\pm$ 0.40
& 0.80 $\pm$ 0.40
& 10.70 $\pm$ 0.08
& 10.70 $\pm$ 0.08 \\

& Action-space (Adaptive)
& 99.20 $\pm$ 0.28
& 0.80 $\pm$ 0.28
& 10.92 $\pm$ 0.12
& 10.92 $\pm$ 0.12 \\

& MPC (Adaptive)
& 100.00 $\pm$ 0.00
& 0.00 $\pm$ 0.00
& 12.75 $\pm$ 0.06
& 12.40 $\pm$ 0.06 \\

\midrule

\multirow{9}{*}{Mixed}

& LoRA ($\kappa_t=0.2$)
& 98.00 $\pm$ 0.00
& 2.00 $\pm$ 0.00
& 14.88 $\pm$ 0.28
& 14.81 $\pm$ 0.28 \\

& LoRA ($\kappa_t=0.6$)
& 94.16 $\pm$ 1.15
& 5.84 $\pm$ 1.15
& 12.66 $\pm$ 0.27
& 12.64 $\pm$ 0.27 \\

& LoRA ($\kappa_t=1.0$)
& 86.08 $\pm$ 2.79
& 13.92 $\pm$ 2.79
& 11.85 $\pm$ 0.18
& 11.84 $\pm$ 0.18 \\

& LoRA (Adaptive-Safe)
& 97.36 $\pm$ 0.67
& 2.64 $\pm$ 0.67
& 14.26 $\pm$ 0.41
& 14.22 $\pm$ 0.40 \\

& LoRA (Adaptive)
& 96.72 $\pm$ 1.21
& 3.28 $\pm$ 1.21
& 13.82 $\pm$ 0.37
& 13.78 $\pm$ 0.38 \\

& LoRA (Adaptive-Efficient)
& 95.28 $\pm$ 1.53
& 4.72 $\pm$ 1.53
& 13.08 $\pm$ 0.33
& 13.06 $\pm$ 0.33 \\

& Full fine-tuning (Adaptive)
& 92.80 $\pm$ 1.20
& 7.20 $\pm$ 1.20
& 12.97 $\pm$ 0.25
& 12.93 $\pm$ 0.24 \\

& Action-space (Adaptive)
& 96.56 $\pm$ 1.22
& 3.44 $\pm$ 1.22
& 13.56 $\pm$ 0.32
& 13.51 $\pm$ 0.32 \\

& MPC (Adaptive)
& 81.36 $\pm$ 1.59
& 18.64 $\pm$ 1.59
& 13.09 $\pm$ 0.08
& 12.69 $\pm$ 0.04 \\

\midrule

\multirow{9}{*}{\shortstack{Spatial\\Clusters}}

& LoRA ($\kappa_t=0.2$)
& 96.00 $\pm$ 0.40
& 4.00 $\pm$ 0.40
& 13.76 $\pm$ 0.40
& 13.71 $\pm$ 0.40 \\

& LoRA ($\kappa_t=0.6$)
& 92.32 $\pm$ 1.73
& 7.68 $\pm$ 1.73
& 12.18 $\pm$ 0.31
& 12.17 $\pm$ 0.30 \\

& LoRA ($\kappa_t=1.0$)
& 88.80 $\pm$ 1.85
& 11.20 $\pm$ 1.85
& 11.61 $\pm$ 0.24
& 11.60 $\pm$ 0.24 \\

& LoRA (Adaptive-Safe)
& 96.08 $\pm$ 1.11
& 3.92 $\pm$ 1.11
& 13.28 $\pm$ 0.32
& 13.26 $\pm$ 0.32 \\

& LoRA (Adaptive)
& 95.68 $\pm$ 0.87
& 4.32 $\pm$ 0.87
& 13.02 $\pm$ 0.30
& 13.00 $\pm$ 0.29 \\

& LoRA (Adaptive-Efficient)
& 95.52 $\pm$ 1.48
& 4.48 $\pm$ 1.48
& 12.90 $\pm$ 0.31
& 12.89 $\pm$ 0.31 \\

& Full fine-tuning (Adaptive)
& 94.56 $\pm$ 1.37
& 5.44 $\pm$ 1.37
& 12.53 $\pm$ 0.27
& 12.50 $\pm$ 0.27 \\

& Action-space (Adaptive)
& 95.84 $\pm$ 1.28
& 4.16 $\pm$ 1.28
& 12.98 $\pm$ 0.28
& 12.94 $\pm$ 0.28 \\

& MPC (Adaptive)
& 86.16 $\pm$ 0.67
& 13.84 $\pm$ 0.67
& 12.94 $\pm$ 0.16
& 12.73 $\pm$ 0.10 \\

\midrule

\multirow{9}{*}{\shortstack{Time-\\varying}}

& LoRA ($\kappa_t=0.2$)
& 98.64 $\pm$ 0.73
& 1.36 $\pm$ 0.73
& 14.85 $\pm$ 0.17
& 14.78 $\pm$ 0.16 \\

& LoRA ($\kappa_t=0.6$)
& 93.36 $\pm$ 1.15
& 6.64 $\pm$ 1.15
& 12.78 $\pm$ 0.32
& 12.76 $\pm$ 0.31 \\

& LoRA ($\kappa_t=1.0$)
& 88.24 $\pm$ 1.19
& 11.76 $\pm$ 1.19
& 11.81 $\pm$ 0.22
& 11.80 $\pm$ 0.22 \\

& LoRA (Adaptive-Safe)
& 97.76 $\pm$ 1.00
& 2.24 $\pm$ 1.00
& 14.35 $\pm$ 0.30
& 14.27 $\pm$ 0.29 \\

& LoRA (Adaptive)
& 96.88 $\pm$ 0.95
& 3.12 $\pm$ 0.95
& 13.78 $\pm$ 0.35
& 13.71 $\pm$ 0.31 \\

& LoRA (Adaptive-Efficient)
& 94.16 $\pm$ 1.22
& 5.84 $\pm$ 1.22
& 12.99 $\pm$ 0.18
& 12.93 $\pm$ 0.18 \\

& Full fine-tuning (Adaptive)
& 92.40 $\pm$ 1.26
& 7.60 $\pm$ 1.26
& 13.02 $\pm$ 0.14
& 12.98 $\pm$ 0.14 \\

& Action-space (Adaptive)
& 96.72 $\pm$ 1.21
& 3.28 $\pm$ 1.21
& 13.74 $\pm$ 0.36
& 13.69 $\pm$ 0.36 \\

& MPC (Adaptive)
& 86.72 $\pm$ 1.04
& 13.28 $\pm$ 1.04
& 12.90 $\pm$ 0.06
& 12.62 $\pm$ 0.04 \\

\bottomrule
\end{tabular}}\vspace{0.1cm}

\end{table*}

\paragraph{Fixed-scale sweep.}

Holding $\kappa_t=0$ throughout a rollout disables the residual and uses the GenSafeNav (Cons.) policy, with the same recurrent-state initialization.
Increasing the fixed scale progressively activates the residual learned from reactive-crowd rollouts, reducing NT and PL in every scenario.
Because the backbone remains frozen, $\kappa_t=1$ denotes the fully activated low-rank residual on the conservative backbone, it is distinct from the separately trained GenSafeNav (Naive) policy.
The sweep reveals that the safety cost of more efficient behavior depends strongly on pedestrian reactivity.
In Non-reactive, increasing the scale from 0 to 1 reduces NT from 16.59 to 12.35 s and PL from 16.52 to 12.33 m, but SR drops from $97.20\%$ to $73.12\%$. In Reactive, the same change reduces NT from 16.24 to 10.87 s and PL from 16.19 to 10.87 m, while SR changes only from $100.00\%$ to $99.20\%$.
Thus, a scale that efficiently exploits reactive pedestrian responses can be unsafe when pedestrians do not respond to the robot.

\paragraph{Bias-controlled adaptation.}
We use $\eta=0.5$, $0.3$, and $0.7$ for Adaptive, Adaptive-Safe, and Adaptive-Efficient, respectively.
Under ground-truth conditioning, 
varying $\eta$ adjusts this trade-off: Adaptive-Safe increases SR at the cost of longer NT and PL, while Adaptive-Efficient favors efficiency with lower SR.
These biased variants are omitted in homogeneous scenarios, where ground-truth conditioning produces effectively the same behavior as neutral adaptation.

\paragraph{Online reactivity-conditioned adaptation.}

In Non-reactive, the adaptive policy remains close to the conservative operating point ($\kappa_t=0$), with nearly unchanged SR ($97.04\%$ versus $97.20\%$), while reducing NT and PL by $3.1\%$.
In Reactive, it matches the fully activated residual ($\kappa_t=1$), achieving the same $99.20\%$ SR, 10.87~s NT, and 10.87~m PL.
In Mixed, the adaptive policy instead operates between these two extremes, maintaining $96.72\%$ SR while reducing NT by $16.8\%$ and PL by $16.7\%$ relative to $\kappa_t=0$.
In contrast, always using $\kappa_t=1$ reduces SR to $86.08\%$.
In Time-varying, oracle adaptation achieves $96.88\%$ SR with $13.7$~m PL, compared with $88.24\%$ SR and $11.8$~m PL at fixed $\kappa=1$, retaining substantially higher success while still improving efficiency over the conservative endpoint.
Together, these results show that \ours adapts the residual strength to the local interaction regime.

\begin{figure}[t]
 \centering
 \includegraphics[width=\linewidth]{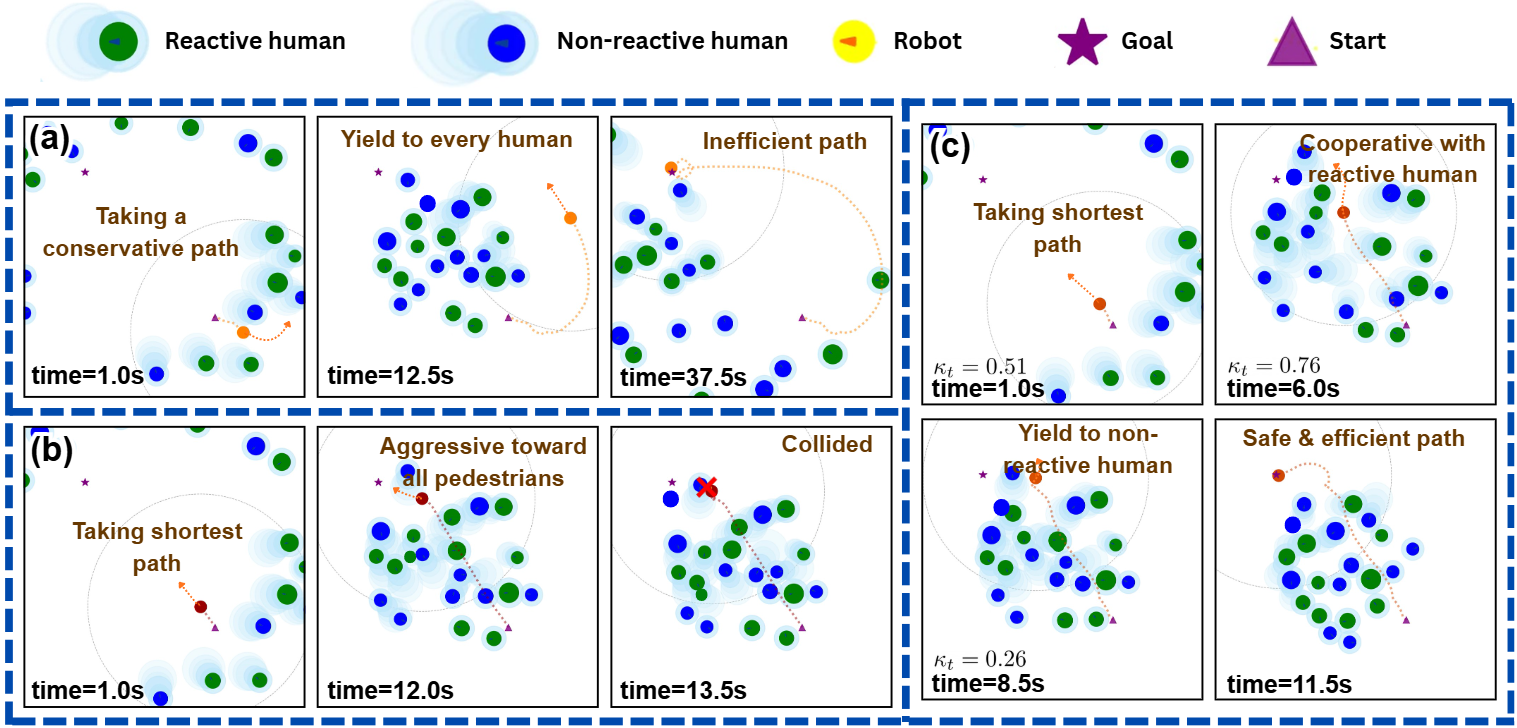}
 \caption{\textbf{Representative mixed-crowd trajectories,} comparing GenSafeNav (Cons.), GenSafeNav (Naive), and \ours.
 (a) The conservative policy treats all pedestrians as non-reactive and takes a long detour.
 (b) The naive policy pursues a shorter route but fails to account for non-reactive pedestrians, leading to collision.
 (c) \ours{} adapts to local robot-directed reactivity, yielding near non-reactive pedestrians while moving more assertively near reactive pedestrians.
 }
 \label{fig:qualitative_revised}
\end{figure}

Fig.~\ref{fig:qualitative_revised} illustrates how \ours{} adapts its navigation behavior within a mixed scenario.
The conservative policy in Fig.~\ref{fig:qualitative_revised}(a) remains safe but detours around the entire crowd, while the naive policy in Fig.~\ref{fig:qualitative_revised}(b) takes a direct route but collides with a non-reactive pedestrian.
In contrast, \ours{} in Fig.~\ref{fig:qualitative_revised}(c) adjusts its behavior online through the adaptation scale $\kappa_t$.
Starting from an intermediate value ($\kappa_t=0.51$), the scale increases to $0.76$ when reactive pedestrians become locally relevant, allowing the robot to take a more direct path, and subsequently decreases to $0.26$ as a non-reactive pedestrian becomes critical, shifting the policy toward the conservative backbone, producing a safe and efficient trajectory within the same episode.

\paragraph{Adaptive MPC.}
\label{sec:adaptive_mpc_ablation}

To evaluate whether reactivity conditioning extends beyond learned policies, we use the adaptation $\kappa_t$ to modulate MPC's clearance around predicted pedestrian trajectories:
\begin{equation}
d^{\min}_{i,t+k\mid t}
=
\hat{\delta}_{i,t+k\mid t}
+
0.5\,\mathrm{m}\,(1-\kappa_t),
\label{eq:adaptive_mpc_clearance}
\end{equation}
where $\hat{\delta}_{i,t+k\mid t}$ is the prediction-uncertainty radius.
Fig.~\ref{fig:adaptation_tradeoff_appendix} shows that MPC also exhibits a monotonic safety--efficiency trade-off, with increasing $\kappa$ reducing both PL and SR.
Table~\ref{tab:lora_ablation_full_revised} shows MPC's adaptive operating point is less safe than the learned adaptive variants in non-reactive and heterogeneous scenarios, indicating that the underlying planner remains important.
Nevertheless, the result shows that reactivity-conditioned adaptation is not limited to the RL policy backbone and can also be incorporated into a model-based planner by modulating its safety constraints online.

\section{Scalability}
\label{sec:adaptation_mechanism_ablation}

We compare the computational scalability of three implementations of the same reactivity-conditioned adaptation rule: LoRA, gated full fine-tuning, and action-space interpolation.
LoRA is particularly attractive because it preserves the conservative backbone and expresses adaptation through a small gated residual, which can better retain the pretrained policy's behavior than
full-parameter adaptation \citep{schmied2023learning}.
Gated full fine-tuning trains all policy parameters from the same conservative initialization on reactive-crowd rollouts and scales the resulting parameter residual:

\begin{equation}
 \Delta\theta_{\mathrm{FT}}
 =\theta_{\mathrm{FT}}-\theta_0,
 \qquad
 \theta_t^{\mathrm{FT}}
 =\theta_0+\kappa_t\Delta\theta_{\mathrm{FT}}.
 \label{eq:full_finetuning_gate}
\end{equation}

Action-space interpolation evaluates the conservative and fully activated LoRA endpoint policies on the same observation and combines their velocity commands, denoted $a_t^{(0)}$ and $a_t^{(1)}$, respectively:

\begin{equation}
 a_t^{\mathrm{interp}}
 =(1-\kappa_t)a_t^{(0)}+\kappa_t a_t^{(1)}.
 \label{eq:action_interpolation}
\end{equation}

\paragraph{Scaling setup.}

We scale the backbone by increasing the hidden-layer width, yielding 2.5M, 25M, 250M, and 500M parameter models. Each model is trained for $2\times10^7$ environment steps on an NVIDIA RTX PRO 6000 Blackwell GPU.
LoRA trains 0.14\%, 0.085\%, 0.030\%, and 0.021\% of the backbone parameter count, respectively, whereas full fine-tuning updates all backbone parameters.
Action-space interpolation reuses the trained LoRA endpoint policies and requires two policy evaluations per control step, executed in parallel in the measured implementation.
P95 denotes the 95th percentile of policy inference latency. 
The latency and peak inference memory measurements concern the policy implementation, excluding perception, trajectory prediction, and reactivity estimation.

\paragraph{Scaling results.}
\begin{table}[t]
    \centering
    \small
    \setlength{\tabcolsep}{4pt}
    \renewcommand{\arraystretch}{0.95}
    \caption{\textbf{Computational scalability.}
    FT denotes full fine-tuning. P95 and memory refer to policy
    inference latency and peak inference memory.
    Action interpolation shares the trained endpoints.}
    \label{tab:scalability_compute}

    \begin{tabular}{@{}l|l|rrr@{}}
        \toprule
        \textbf{Backbone} & \textbf{Method}
        & \textbf{Train (h)} & \textbf{P95 (ms)}
        & \textbf{Memory (MB)} \\
        \midrule
        \multirow{3}{*}{2.5M}
        & LoRA     & 6.30   & 1.90 & 47.2 \\
        & Gated FT       & 6.44   & 2.20 & 56.6 \\
        & Action-interpolation & Shared & 3.06 & 89.7 \\
        \midrule
        \multirow{3}{*}{25M}
        & LoRA     & 6.80   & 1.97 & 287 \\
        & Gated FT       & 7.58   & 2.23 & 398 \\
        & Action-interpolation & Shared & 3.03 & 510 \\
        \midrule
        \multirow{3}{*}{250M}
        & LoRA     & 9.26   & 2.52 & 2774 \\
        & Gated FT       & 15.92   & 3.20 & 3784 \\
        & Action-interpolation & Shared & 4.05 & 4688 \\
        \midrule
        \multirow{3}{*}{500M}
        & LoRA     & 9.32   & 3.18 & 5614 \\
        & Gated FT       & 22.52  & 4.16 & 7641 \\
        & Action-interpolation & Shared & 5.63 & 9459 \\
        \bottomrule
    \end{tabular}%
\end{table}

Table~\ref{tab:scalability_compute} shows that LoRA and full fine-tuning have similar training times at 2.5M parameters (6.30 versus 6.44 h), while their costs diverge as the backbone grows.
At 500M, LoRA requires 9.32 h compared with 22.52 h for full fine-tuning, corresponding to a $2.42\times$ training speedup.
Its P95 inference latency is 3.18 ms versus 4.16 ms, and peak inference memory is 5,614 MB versus 7,641 MB. 
Action-space interpolation requires 5.63 ms and 9,459 MB at the same scale.
For a square layer of width $d$, a full parameter residual contains $d^2$ parameters, whereas rank-$r$ factors contain $2dr$.
At fixed rank, their relative parameter cost decreases with width.
Together, these measurements show the computational advantage of the low-rank implementation for larger backbones.
These are implementation-specific resource measurements. 
Frozen-backbone computation, rollout collection, and residual evaluation still incur costs, so trainable-parameter savings do not imply proportional speedups. 
Fig.~\ref{fig:adaptation_tradeoff_appendix} shows similar safety--efficiency trends for LoRA and gated full fine-tuning at 2.5M parameters under ground-truth reactivity conditioning, although their operating points differ.
The larger backbones exhibit navigation performance similar to their respective 2.5M counterparts, these results are omitted for brevity.
These results suggest that the behavioral benefit of reactivity-conditioned adaptation is not specific to LoRA.



\end{document}